\documentclass[letterpaper]{article} %
\usepackage[draft]{aaai25}  %
\usepackage{times}  %
\usepackage{helvet}  %
\usepackage{courier}  %
\usepackage[hyphens]{url}  %
\usepackage{graphicx} %
\usepackage{natbib}  %
\usepackage{caption} %
\usepackage{algorithm}
\usepackage{algorithmic}
\usepackage{amsmath}
\usepackage{amssymb}
\usepackage{booktabs}
\usepackage{colortbl}
\usepackage{newfloat}
\usepackage{listings}
\DeclareCaptionStyle{ruled}{labelfont=normalfont,labelsep=colon,strut=off} %
\floatstyle{ruled}
\newfloat{listing}{tb}{lst}{}
\floatname{listing}{Listing}
\usepackage{xcolor}
\definecolor{lightgray}{rgb}{0.95, 0.95, 0.95}
\title{LIMP: Large Language Model Enhanced Intent-aware Mobility Prediction}

\author{Songwei Li, Jie Feng, Jiawei Chi, Xinyuan Hu$^{*}$, Xiaomeng Zhao, Fengli Xu  \\
  Department of Electronic Engineering, Tsinghua University, Beijing, China, \\
  $^{*}$Quantitative Method and Theory, Emory University, USA, \\
  \{fengjie,fenglixu\}@tsinghua.edu.cn
  }

\begin{document}
\maketitle

\begin{abstract}
Human mobility prediction is essential for applications like urban planning and transportation management, yet it remains challenging due to the complex, often implicit, intentions behind human behavior. Existing models predominantly focus on spatiotemporal patterns, paying less attention to the underlying intentions that govern movements. Recent advancements in large language models (LLMs) offer a promising alternative research angle for integrating commonsense reasoning into mobility prediction. However, it is a non-trivial problem because LLMs are not natively built for mobility intention inference, and they also face scalability issues and integration difficulties with spatiotemporal models. To address these challenges, we propose a novel LIMP (\underline{L}LMs for \underline{I}ntent-ware \underline{M}obility \underline{P}rediction) framework. Specifically, LIMP introduces an ``Analyze-Abstract-Infer'' (A2I) agentic workflow to unleash LLM’s commonsense reasoning power for mobility intention inference. Besides, we design an efficient fine-tuning scheme to transfer reasoning power from commercial LLM to smaller-scale, open-source language model, ensuring LIMP’s scalability to millions of mobility records. Moreover, we propose a transformer-based intention-aware mobility prediction model to effectively harness the intention inference ability of LLM. Evaluated on two real-world datasets, LIMP significantly outperforms baseline models, demonstrating improved accuracy in next-location prediction and effective intention inference. The interpretability of intention-aware mobility prediction highlights our LIMP framework’s potential for real-world applications. Codes and data can be found in \url{https://github.com/tsinghua-fib-lab/LIMP}.
\end{abstract}

\section{Introduction} \label{sec:intro}
Predicting human mobility behavior is a crucial task with significant implications for various domains, including urban planning, transportation management, and public safety. However, the inherent complexity of human mobility poses substantial challenges, especially the implicit intentions that are often not directly observable. Previous studies have shown that human researchers can infer the intention of human movements with high accuracy by examining their spatiotemporal trajectory~\cite{jiang2016timegeo,liccardi2016know}. However, it is not scalable to ask human researchers to manually label mobility data. Thus, most of existing mobility prediction models~\cite{liu2016predicting, feng2018deepmove, sun2020go, luo2021stan, yang2022getnext} focus on capturing spatiotemporal patterns using advanced recurrent network and attention models. While these methods have shown promise, they often fail to effectively model the underlying intentions that drive each movement. This limitation highlights the need for new approaches that can incorporate a deeper understanding of human behavior.

Recent advancements in large language models (LLMs) have demonstrated emergent capabilities in commonsense reasoning~\cite{wei2022emergent,wei2022chain}, offering a novel research angle for intention-aware mobility prediction. Despite this promise, several challenges remain in leveraging LLMs for mobility prediction. First, LLMs are not inherently optimized for inferring behavioral intentions from spatiotemporal data. Directly prompting LLMs to interpret human movements often yields inaccurate or overly generalized results, limiting their utility in mobility prediction. Second, the massive size and proprietary nature of state-of-the-art LLMs, such as GPT-4~\cite{achiam2023gpt}, present practical challenges, including high API costs and the inability to deploy these models locally. These factors hinder the scalability of LLM-based mobility prediction in real-world applications. Third, the domain-specific nature of spatiotemporal deep learning models and LLMs creates a disconnect, making it unclear how to effectively integrate the two to enhance prediction accuracy.

In response to these challenges, we propose a novel framework, LIMP (\underline{L}LMs for \underline{I}ntent-ware \underline{M}obility \underline{P}rediction), designed to harness the commonsense reasoning abilities of LLMs for intention-aware mobility prediction. The framework comprises three key components. First, we introduce an ``Analyze-Abstract-Infer" (A2I) agentic workflow that guides LLMs through the process of mobility intention inference in a principal manner, which emulates the methodology of human expert annotators. The A2I workflow enables LLMs to reason through the intentions behind movements step-by-step: analyzing notable features, abstracting high-level insights, and inferring the most likely intention with a comprehensive assessment. Second, we propose an efficient fine-tuning scheme that transfers the reasoning power of large commercial LLMs, such as GPT-4, to smaller, locally deployable models like Llama 3~\cite{dubey2024llama}. This approach ensures that our framework can scale to handle millions of mobility records at low cost. Finally, we design a transformer-based intention-aware mobility prediction model that effectively integrates the inferred intentions from LLMs to enhance next-location prediction.

Our framework is evaluated on two real-world datasets, demonstrating significant improvements over baseline models. Specifically, LIMP achieves a 6.64\% to 9.52\% increase in top-1 accuracy for next-location prediction compared to state-of-the-art baselines. Additionally, the A2I workflow boosts LLMs' mobility intention inference accuracy by 16.28\%. The fine-tuning scheme successfully transfers the reasoning abilities of GPT-4 to Llama 3 without any statistically significant performance loss. Ablation studies further validate the effectiveness of our design choices. 

In summary, our contributions are threefold. 
\begin{itemize}
    \item We introduce a novel framework, LIMP, that leverages the commonsense reasoning power of LLMs for enhanced mobility prediction, incorporating intention inference to improve both performance and interpretability.
    \item We develop a fine-tuning strategy that enables the deployment of high-performing, cost-effective, and scalable models by transferring capabilities from large, proprietary LLMs to smaller, open-source alternatives.
    \item We provide extensive empirical validation on real-world datasets, demonstrating the robustness and applicability of our approach to practical scenarios.
\end{itemize}

\section{Method} \label{sec:method}
\subsection{Overview}
\begin{figure*}
    \centering
    \includegraphics[width=1.\textwidth]{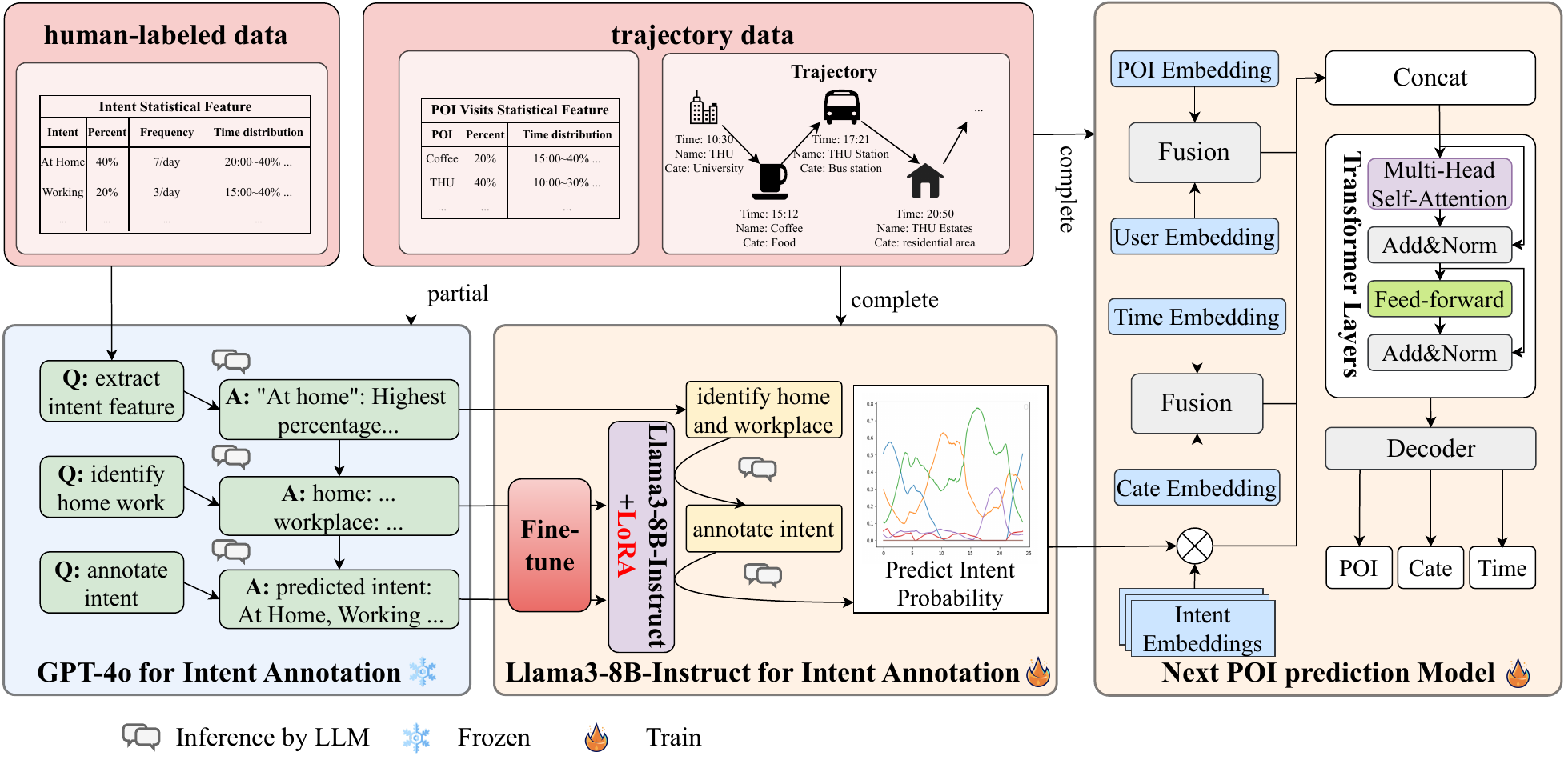}
    \caption{The framework of LIMP, including Analyze-Abstract-Infer (A2I) agentic workflow for intent annotation, agentic workflow fine-tuning schema, and transformer based mobility prediction model.}
    \label{figs:framework}
\end{figure*}

As Figure~\ref{figs:framework} shows, LIMP contains three main steps ranging from GPT-4o based A2I agentic workflow for zero-shot intent annotation for mobility trajectory, agentic workflow fine-tuning framework for distilling the capabilities from GPT-4o to smaller LLMs (e.g., Llama3-8B), and intent enhanced mobility prediction model. Detailed design of three steps are introduce in the following sections.

\subsection{Intent Annotation via A2I Agentic Workflow}
In this section, we present a novel agentic workflow specifically developed to leverage the commonsense reasoning capabilities of LLMs, such as GPT-4o, for inferring user intentions from mobility data. Recognizing that LLMs inherently lack specialized capabilities for mobility intention inference, we proposed the \textbf{Analyze-Abstract-Infer (A2I) Agentic Workflow}, a systematic approach inspired by methodologies employed by human researchers for annotating trajectories and designing algorithmic processes~\cite{zeng2017visualizing, chen2021identifying, liccardi2016know, jiang2016timegeo}. We designed A2I with three sequential steps following the idea of Chain-of-Thought (CoT)~\cite{wei2022cot}, intent feature analysis, summative insights generation and user intentions inference. Detailed designs of them are introduce below.

\subsubsection{Expert-Knowledge Guided Intent Feature Analysis}
Our approach began with analyzing the expert-knowledge statistics derived from a small, human-labeled dataset.  Inspired by the TimeGeo modeling framework for urban mobility \cite{jiang2016timegeo}, we identified features that are indicative of user behavior patterns and essential for guiding LLMs in the task of mobility intention inference by providing necessary expert knowledge. The features analyzed include:
\begin{itemize}
    \item Percentage Distribution: We computed the relative frequency of specific intents—\textit{'At Home', 'Working', 'Running Errands'}—to capture the overall distribution of user behaviors in the dataset.
    \item Average Visit Frequency: This metric quantified the frequency with which users visited specific Points of Interest (POIs), providing insights into habitual behaviors and their associated intents.
    \item Time Distribution: For each of the six intents—\textit{'At Home', 'Working', 'Running Errands', 'Eating Out', 'Leisure and Entertainment', 'Shopping'}—we calculated the temporal distribution of visits, allowing us to model typical user behavior patterns across different times of the day.
\end{itemize}
These knowledge served as the foundation for generating summative insights.

\subsubsection{Summative Insights Generation}
Human researchers are adept at synthesizing statistical data into high-level insights that can be generalized across different scenarios and tasks \cite{zhao2024expel}. Emulating this capability, we leveraged the commonsense reasoning abilities of LLMs to generate \textit{n} summative insights from the previously identified features that could aid LLMs in differentiating user intents in the subsequent process. For instance, the LLM was tasked with identifying key patterns, such as the peak activity times for different intents or the typical working hours for users.
Then, we transformed the larger dataset's raw mobility data into a structured format suitable for intent prediction, which: first divided the trajectory data into daily segments for each user, capturing the sequence of POIs visited along with key metadata; and then computed the visit frequency and time distribution for each POI, establishing a statistical basis for context-aware prediction.

\subsubsection{User Intentions Inference}
Finally, we used in-context learning (ICL) to synthesize the insights derived from feature extraction and statistical analysis into a cohesive intent inference process:
\begin{itemize}
    \item Home and Workplace Identification (HWI): Leveraging the high-level insights generated from \textit{intent feature analysis}, combined with the POI statistics from \textit{summative insights generation} into the prompt, we applied ICL to guide GPT-4o in predicting the locations most likely corresponding to the user's home and workplace. This initial prediction step was conducted once per user at the outset of the prediction process.
    \item Daily Intent Prediction: For each stay within the user's daily trajectory, GPT-4o predicted the user's intent by considering the identified home and workplace locations, the sequence of POIs, and the time of day. By using ICL, the model could apply its understanding of typical behaviors to infer intents in novel situations.
\end{itemize}
By first identifying the home and workplace, and instructing the LLM to maintain consistency with these predicted locations, we streamlined the analysis process and improved the accuracy of subsequent predictions.

\subsection{Agentic Workflow Fine-tuning}
In this section, we annotate the entire dataset with intents and predict the probability distribution of the next movement's intent based on historical data.
For large datasets, using high-parameter models such as GPT-4o for intentions inference incurs excessive costs, making it inefficient to annotate the entire dataset directly. On the other hand, smaller models like Llama3-8B perform poorly on related tasks and fail to meet the expected performance. To address this issue, we employed a smaller model, Llama3-8B-Instruct, and fine-tuned it with data generated by GPT-4o. This approach allows the model to achieve performance on user intentions inference tasks that is close to that of GPT-4o. The prompts of fine-tune are similar to the user intentions inference part of A2I workflow. We only modified some parts to better fine-tune it. The fine-tuning process consists of two tasks.
\begin{itemize}
\item  \textbf{Task 1}: Based on the distribution of visit times to POIs by specific users and the summative insights of the human-labeled dataset extracted by GPT-4o, determine the POIs of the users' homes and workplaces.
\item \textbf{Task 2}: For a specific user, given the POIs of their home and workplace, annotate the intent for each movement in a segment of their trajectory.
\end{itemize}

We extracted 100 users from training sets in both Beijing and Shanghai, using 20\% of each user's trajectory information. This data was annotated by GPT-4o and then used to fine-tune the \textit{llama3-8B-Instruct} model, enabling the \textit{llama3-8B} model to perform both tasks simultaneously. For detailed fine-tuning prompts and answers, please see the appendix.

Through experiments, we found that providing the probability distribution of the next movement's intent based on historical data is more effective than directly giving the intent information. We developed a method based on expert knowledge to calculate the intent probability distribution.

We believe that people tend to have the same behavioral patterns at the same time each day, and thus the historical intent distribution can be used to infer future intent distribution. Specifically, for the intent sequence of a user \(u\) corresponding to all of his movements in the train dataset, \(( Q_1, Q_2, Q_3, \ldots Q_N )\), and their corresponding time sequence \(( t_1, t_2, t_3, \ldots t_N )\), we consider that an intent \( Q_i \) recorded over a period provides a likelihood of this intent occurring. The influence period is defined as \( t_{begin,i} = \max(t_{i-1}, t_i-T) \) and \( t_{end,i} = \min(t_{i+1}, t_i+T) \), where \( T \) is a parameter representing the maximum influence time range.

For each intent \( I_j \), we construct the function \( f_{I_j}(t) \) as follows to represent the effect of the intent:
\[ f_{I_j}(t) = \sum_{Q_i = I_j} \max \left(0, \min \left( \frac{t - t_{begin,i}}{t_i - t_{begin,i}}, \frac{t_{end,i} - t}{t_{end,i} - t_i} \right) \right) \]
For a specific time \( t_0 \) within a day, the predicted probability of intent \( I_j \) occurring is calculated as:
\[ P(I_j \mid t_0, u) = \frac{\sum_k f_{I_j}(t_0 + k\Delta t)}{\sum_{k, j} f_{I_j}(t_0 + k\Delta t)} \]
where \( \Delta t \) represents a time interval of a day.\\

\subsection{Intent Enhanced Mobility Prediction}
In this section, we present a mobility prediction model that effectively utilizes intent information along with other trajectory data to achieve high prediction accuracy. Among the numerous previous approaches, models employing the transformer architecture have demonstrated outstanding performance. Therefore, our model considers using the transformer architecture. Specifically, we draw inspiration from the GETNext~\cite{yang2022getnext} model to develop our own model. GETNext is an excellent mobility prediction model that leverages the transformer architecture to fully utilize temporal, user, POI, POI category and map information. Building on this model, we incorporate intent information and simplify the original model structure, achieving promising results.

Our model's input sequence unit is defined as \((u, p, c, t, P(I|t, u))\), where \(u\) is the user ID, \(p\) is the current POI of the user, \(c\) is the category of the POI, \(t\) is the time of day of the next movement, and \(P(I|t, u)\) is the predicted probability distribution of the next movement's intent. To effectively handle the predicted probability information, we used a method of embedding intents weighted by their probabilities. This means that the higher the probability of a certain intent, the closer the overall intent embedding is to the individual embedding of that intent. Compared to providing only the single most likely intent information, this method offers richer intent probability information. This makes it easier for the model to determine the extent to which it should rely on the intent information, resulting in a more effective combination of intent information and trajectory information.
Specifically, for each intent \(I_i\), we map it to a high-dimensional vector 
\[e_{I_i} = f_{embed}(I_i) \in \mathbb{R}^\Gamma\]
The embedding of the total intent at time \(t\) is defined as
\[e_I = \sum_i e_{I_i} P(I_i|t, u)\]
We simplified the GETNext model by omitting the generic movement learning part and directly embedding  \(p\):  
\[e_p = f_{embed}(p) \in \mathbb{R}^\Omega\]
The definitions of the embeddings for \(u\), \(c\), and \(t\), denoted as \(e_u\), \(e_c\), and \(e_t\), and the fused embeddings \( e_{p,u}\), \(e_{c,t}\) are consistent with GETNext.
The final embedding representation for each movement is \(e_q = [ e_{p,u}, e_{c,t}, e_I ]\). For a trajectory \((q_1, q_2, \ldots, q_N)\), we use \((e_{q_1}, e_{q_2}, \ldots, e_{q_N})\) as the input to the transformer Encoder. We adopted the same Transformer Encoder, MLP Decoders structure and loss function as GETNext. 

\section{Experiment} \label{sec:exp}

\subsection{Settings}
\subsubsection{Dataset}
To evaluate our model, we collected two representative datasets: one comprising mobile application location data from a popular social network vendor, referred to as the Beijing dataset, and the other consisting of call detail records (CDR) data from a major cellular network operator, referred to as the Shanghai dataset. These datasets are named after the cities from which the data was sourced: Beijing and Shanghai. The data generation mechanisms differ significantly: mobile application data with location records is generated on application servers when users request location-based services within the app, such as searches, check-ins, and similar activities, while call detail records data with location information is generated at cellular network base stations when users access the network for communication or data services. 

 The Beijing dataset includes check-ins from late September 2019 to late November 2019, spanning approximately three months, while Shanghai dataset primarily covers January 2016. The number of users, locations, and check-ins for each dataset are detailed in Table~\ref{tab:data}. Following the common practice of mobility prediction~\cite{feng2018deepmove, sun2020go, yang2022getnext}, we segmented trajectories into fixed-length sessions and applied a sliding window over the dataset to make full use of the data during training. Specifically, for a user with $m$ check-ins, and a fixed length $n$, the processed trajectories will contain $(m-n+1)  n$ check-ins, where each sequence of $n$ check-ins forms a trajectory, and consecutive trajectories overlap by $n-1$ point. This strategy is also applied in all the baselines to ensure the fair comparison.
\setlength{\tabcolsep}{1mm}
\begin{table}[ht]
    \centering
    \begin{tabular}{lcccc} \toprule
         &  \textbf{Duration} &  \textbf{Users} &  \textbf{POIs} & \textbf{Records} \\ \midrule 
        \textbf{Beijing} &  3 months &  1566 &  5919 & 744813 \\ 
        \textbf{Shanghai} & 1 month &  841 &  6955 & 215379 \\ \bottomrule
    \end{tabular}
    \caption{Basic statistics of two mobility datasets.}
    \label{tab:data}
\end{table}

\subsubsection{Manual Intent Annotation on Real Data}
We hire 10 undergraduates to manually annotated intent labels for a small dataset. The annotation process was conducted as follows:

\textbf{Identification of Home and Work Locations}:  The primary step involves identifying the home and work locations as anchor points. This is determined based on the frequency, time, and location of Points of Interest (POIs) within a user's trajectory. Specifically, if a user frequently visits a location during the night until the next morning, and this location is in a residential area, school, or other dwelling place, we consider the location as home.  Conversely, if a user consistently visits a location during typical work hours, the location will be considered as working place. The intents for these anchor points are accordingly set as "At Home" and "Working".

\textbf{Determination of Other Trajectory Intents}: For other trajectories, intent is assigned based on the POI type and the timing of the visit. If a POI is a dining location, such as a restaurant or food street, and the visit occurs during meal times, the intent is labeled as "Eating Out." If the POI corresponds to recreational venues like bars, game centers, or sports grounds, the intent is marked as "Leisure and Entertainment". If the POI is located at a store or shopping mall, the intent is labeled as "Shopping". Trajectories that do not fit into the aforementioned categories, such as medical visits, or those where the POI cannot be precisely identified on the map, are labeled with the intent "Running Errands".

\subsubsection{Baselines}
We consider the following methods as baselines to benchmark the performance of our model. 
\begin{itemize}
    \item \textbf{RNN~\cite{graves2012supervised}}: a classical model for processing sequential data, capturing temporal dependencies through recurrent connections.
    \item \textbf{DeepMove~\cite{feng2018deepmove}}: a model that combines recurrent networks and attention layer to capture multi-scale temporal periodicity of human mobility. 
    \item \textbf{STAN~\cite{luo2021stan}}: a model that utilizes a dual-attention structure to enhance next-location recommendation by aggregating spatio-temporal correlations and incorporating personalized item frequency (PIF).
    \item \textbf{LSTPM~\cite{sun2020go}}: it integrates a non-local network for capturing long-term preferences and a geo-dilated recurrent neural network for short-term preferences modelling.
    \item \textbf{Graph-Flashback~\cite{rao2022graph}}: it enhances point-of-interest (POI) representation by combining a graph convolution network (GCN) with a POI transition graph.
    \item \textbf{GETNext~\cite{yang2022getnext}}: a state-of-art model that introduces a user-agnostic global trajectory flow map and a novel graph enhanced transformer to improve next POI recommendation.
\end{itemize}

\subsubsection{Training Settings}
For baselines, we use official codes from authors to implement RNN, DeepMove~\footnote{https://github.com/vonfeng/DeepMove}, STAN~\footnote{https://github.com/yingtaoluo/Spatial-Temporal-Attention-Network-for-POI-Recommendation}, LSTPM~\footnote{https://github.com/NLPWM-WHU/LSTPM}, Graph-Flashback~\footnote{https://github.com/kevin-xuan/Graph-Flashback}, GETNext~\footnote{https://github.com/songyangme/GETNext}. 
The hyperparameters for the models were adjusted according to the dataset, beginning with the default settings provided in the original codebases. For both RNN and DeepMove, the embedding dimensions for location and user were set to 128, with the Adam optimizer applied using a learning rate of 1e-3. The maximum number of training epochs was set to 20. For LSTPM, the embedding dimensions for location and user were set to 50, with the Adam optimizer configured with a learning rate of 1e-4, and a maximum of 20 training epochs. Similarly, STAN utilized embedding dimensions of 50 for both location and user, an Adam optimizer with a learning rate of 3e-3, and a maximum of 20 training epochs. Graph-Flashback employed embedding dimensions of 100 for both location and user, with the Adam optimizer set to a learning rate of 1e-2, and a maximum of 100 training epochs. Lastly, GETNext was configured with embedding dimensions of 128 for both location and user, an Adam optimizer with a learning rate of 1e-3, and a maximum of 200 training epochs. The length of all trajectories was standardized to 12. All other unique parameters for each model, as well as those not explicitly mentioned, adhered to the default settings of the respective codebases.

We fine-tuned the llama3-8B-Instruct model using LLaMA-Factory~\cite{Zheng2024LlamaFactoryUE}. Our model is deployed on Google Colab, utilizing cloud computing resources for fine-tuning. The original model was a 4-bit quantized version of llama3-8B-Instruct. We utilized LoRA adapters and performed fine-tuning using 4-bit QLoRA. The training was conducted for 3 epochs and a batch size of 2. The parameter of learning rate is set to 5e-5. After fine-tuning, we merged the LoRA adapter weights with the non-quantized llama3-8B-Instruct to obtain the final model. We used the PyTorch framework for training the transformer based mobility prediciton model on the following hardware platform (CPU: Intel Xeon Platinum 8358, GPU: NVIDIA GeForce RTX 4090). The embedding dimensions for POI and user are \(\Omega = 128\), while the embedding dimensions for time, POI category \(\Psi\), and intent \(\Gamma\) are all 32. The transformer architecture includes two encoder layers. In the encoder layers, the feed-forward network dimension is 1024, and the multi-head attention module uses two attention heads. The transformer architecture employs a dropout with a probability of 0.3. We used the Adam optimizer with a learning rate of 1e-3, a weight decay rate of 5e-4 and a maximum of 200 training epochs.
\subsubsection{Metrics}
We used two common metrics in  existing works~\cite{sun2020go, luo2021stan} to evaluate the performance of our model: Accuracy at k (Acc@k) and Mean Reciprocal Rank at 5 (MRR@5). Acc@k measures the proportion of the true next POI is found among the top k recommendations. MRR@5 evaluates the average of the reciprocal ranks of the true next POI in the top 5 recommendations. Define

\[
\text{Acc@k} = \frac{1}{N} \sum_{i=1}^{N} \mathbb{I}(r \leq k)
\]
\[
\text{MRR@5} = \frac{1}{N} \sum_{i=1}^{N} \frac{1}{R}
\]
where \( N \) is the total number of queries, \( \mathbb{I}(\cdot) \) is the indicator function that returns 1 if the condition is true and 0 otherwise, \(r\) is the rank position of the true next POI. \(R\) is equal to \(r\) only when \(r \leq 5\); otherwise, it tends to infinity.\\

\subsection{Main Results}
The main results on the Beijing and Shanghai datasets are shown in Table \ref{table:beijing} and Table \ref{table:shanghai}. Our method outperforms all baseline methods across various metrics on both datasets. For instance, in terms of Acc@1 on the Beijing dataset, the best baseline achieves 0.4547, while our method reaches 0.4980, representing an approximate 10\% relative improvement. Similarly, our method also shows nearly 10\% relative improvements over the best baseline method in terms of Acc@10 and MRR@5 on the Shanghai dataset. Our method performed worse on the Shanghai dataset compared to the Beijing dataset. This could be because the model architecture is less effective with the smaller data volume and higher total number of POIs in the Shanghai dataset. As a comparison, the GETNext model, which has a very similar architecture to ours, also showed a significant performance drop on the Shanghai dataset compared to the Beijing dataset.

\begin{table}[ht]
\centering
\begin{tabular}{lcccc} 
\toprule
\textbf{Models} & \textbf{Acc@1} & \textbf{Acc@5} & \textbf{Acc@10} & \textbf{MRR@5} \\ 
\midrule
\textbf{RNN}  &  0.2290&  0.3667&  0.3941&  0.2846\\ 
\textbf{DeepMove}  &  0.3129&  0.5202&  0.5482&  0.3999\\ 
\textbf{STAN}  &  0.3270&  0.6532&  0.7419&  0.4548\\ 
\textbf{LSTPM}  &  0.4291&  0.7910&  0.8202&  0.5826\\ 
\textbf{Graph-Flashback}  &  0.4387&  0.8103&  0.8508&  0.5923\\ 
\textbf{GETNext} &  \underline{0.4547}&  \underline{0.8175}&  \underline{0.8596}&  \underline{0.6126}\\ 
\midrule
\textbf{Ours} &  \textbf{0.4980}&  \textbf{0.8337}&  \textbf{0.8718}&  \textbf{0.6444}\\ 
\textbf{$\delta$}&  9.52\%&  1.98\%& 1.42\%&  5.19\%\\
\bottomrule
\end{tabular}
\caption{Intent-aware mobility prediction results on Beijing data.}
\vspace{-0.5cm}
\label{table:beijing}
\end{table}

\begin{table}[ht]
\centering
\begin{tabular}{lcccc} 
\toprule
\textbf{Models} & \textbf{Acc@1} & \textbf{Acc@5} & \textbf{Acc@10} & \textbf{MRR@5} \\ 
\midrule
\textbf{RNN} &  0.2530&  0.3899&  0.4232&  0.3084\\ 
\textbf{DeepMove} &  0.2713&  0.3893&  0.4123&  0.3199\\ 
\textbf{STAN} &  0.2566&  0.5411&  0.6544&  0.3641\\ 
\textbf{LSTPM} &  \underline{0.4489}&  \underline{0.7018}&  0.7422&  \underline{0.5518}\\ 
\textbf{Graph-Flashback} &  0.4351&  0.6966&  \underline{0.7445}&  0.5416\\ 
\textbf{GETNext} &  0.4177&  0.6782&  0.7363&  0.5348\\ 
\midrule
\textbf{Ours} &  \textbf{0.4787}&  \textbf{0.7517}&  \textbf{0.8154}&  \textbf{0.6025}\\ 
\textbf{$\delta$} &  6.64\%&  7.11\%& 9.52\%&  9.19\%\\
\bottomrule
\end{tabular}
\caption{Intent-aware mobility prediction results on Shanghai dataset.}
\vspace{-0.5cm}
\label{table:shanghai}
\end{table}

\subsection{Intent Prediction Analysis}
We evaluated our intent prediction workflow with multiple experimental setups on the small, human-labeled dataset to explore how each component in the workflow affects the performance. Results in Table \ref{tab:prediction_metrics} show that our full workflow achieved the highest accuracy, outperforming both zero-shot prediction and workflow without generating features or home and work identification.

\begin{figure}
    \centering
    \includegraphics[width=1\linewidth]{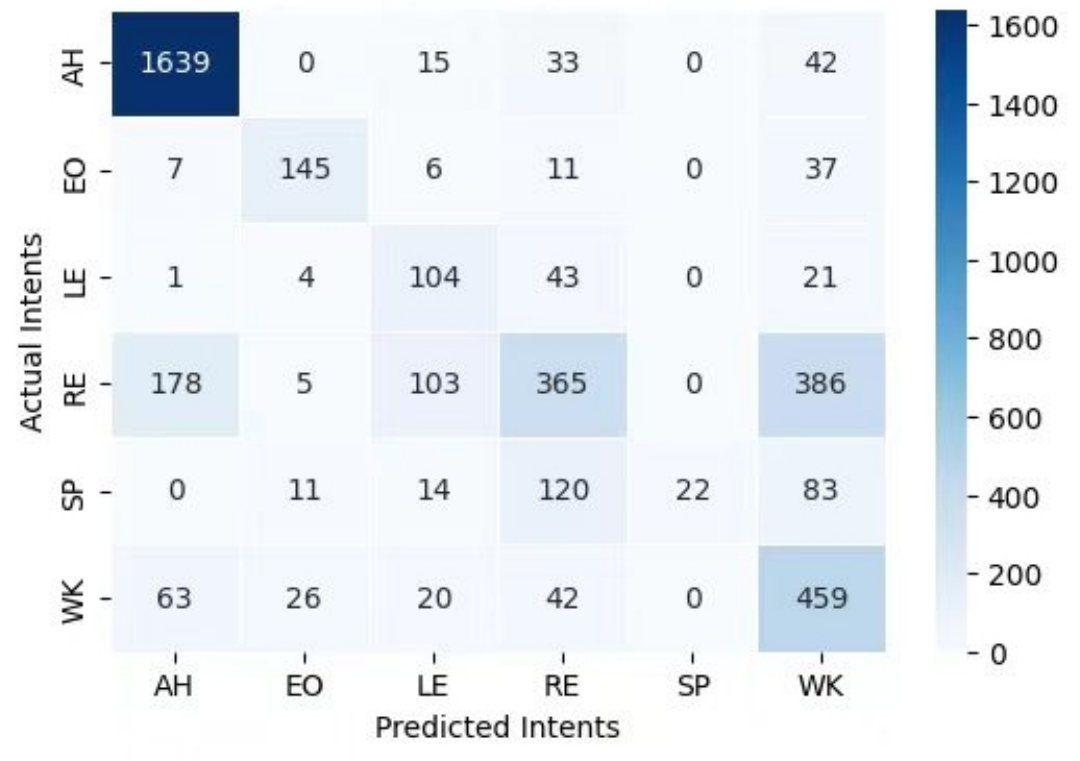}
    \caption{Confusion matrices of intents. AH, EO, LE, RE, SP and WK refer to "At Home", "Eating Out", "Leisure and Entertainment", "Running Errands", "Shopping" and "Working" respectively.
    }
    \vspace{-0.5cm}
    \label{figs:tab4}
\end{figure}

The results indicate that the step of home and workplace identification is a critical factor in improving prediction accuracy. Specifically, the workflow with home and workplace identification demonstrated a substantial increase in accuracy compared to the workflow without this step. In contrast, the generation of additional features did not significantly enhance performance when compared to the base workflow. We analyzed the annotation of specific intents by compiling confusion matrices for six types of intents on human-labeled dataset, as shown in Figure \ref{figs:tab4}. We found that the model's predictive ability is excellent for the intents of At home, Working, Eating Out, and Leisure and Entertainment. Among the incorrect annotations, the most common error is labeling Running errands as Working. This is because the POIs associated with these two intents have similar attributes, and the timing of these activities often overlaps, making it challenging to perfectly distinguish between them. Additionally, Shopping is frequently mislabeled as other intents because the POIs where shopping occurs often serve multiple functions beyond just shopping.

\begin{table}[ht]
\centering
\begin{tabular}{ccccc}
\toprule
\textbf{Experiments} & \textbf{Accuracy} & \textbf{Precision} & \textbf{Recall} & \textbf{F1 Score} \\
\midrule
\textbf{A2I Workflow} & \textbf{67.9\%} & \textbf{0.715} & \textbf{0.679} & \textbf{0.657}   \\ 
\textbf{NFE}          & 67.2\%  & 0.707  & 0.672  & 0.651 \\ 
\textbf{NHWI}         & 60.0\%  & 0.627  & 0.600  & 0.575 \\ 
\textbf{ZS}    & 59.0\%  & 0.609  & 0.590  & 0.565 \\ 
\bottomrule
\end{tabular}
\caption{Ablation study of the prompt designs in A2I workflow, where `NFE' refers to the model run without feature engineering, and `NHWI' refers to the model run without home-work identification, `ZS' denotes the basic prompt settings.}
\vspace{-0.5cm}
\label{tab:prediction_metrics}
\end{table}

\subsection{Ablation Study}
We conducted ablation experiments on the Beijing dataset to demonstrate the effectiveness of our designed architecture. There are four experiments: 
\begin{itemize}
\item \textbf{LIMP}: The experiment with the complete model architecture.
\item \textbf{Max-Prob}: This experiment does not use probabilities to weight the embeddings; instead, it only uses the embedding of the intent with the highest probability, aiming to verify the effect of probability weighting on intent embeddings.
\item \textbf{Train-Real}: This experiment does not predict the intent for the training set but uses the next intent generated by LLM as the predicted intent to generate embeddings. For the test and validation sets, the embeddings of the intent with the highest probability are used. This experiment compares the effects of training the model with real intent and predicted intent.
\item \textbf{w/o intent}: The model without the intent input.
\end{itemize}

The experimental results are shown in Table \ref{tab:Ablation Study}. Overall, the Full Model achieved the best performance. The Max-Prob model, which does not provide the probability distribution information of the intent, and the w/o intent model, which does not provide any intent information, both showed performance degradation compared to the Full Model. The Train-Real experiment had a relatively high Acc@1 indicator but performed significantly worse than the other three models in terms of Acc@5, Acc@10, and MRR@5 indicators, making its overall performance unsatisfactory. This is because the training set directly input the real intent, bypassing the intent prediction step, causing the model to rely too heavily on the intent and become insensitive to other input information. During testing, the intent can only be predicted based on historical data, reducing the success rate of direct intent prediction and thus leading to performance degradation.

\begin{table}[ht]
    \centering
    \begin{tabular}{ccccc}
    \toprule
    \textbf{Experiments} & \textbf{Acc@1} & \textbf{Acc@5} & \textbf{Acc@10} & \textbf{MRR@5} \\
    \midrule
    Full Model & 0.4980& 0.8337& 0.8718& 0.6444\\
     Max-Prob& 0.4835& 0.8328& 0.8726&0.6356\\
     Train-Real& 0.4989& 0.7623& 0.7916&0.6112\\ 
    w/o intent & 0.4670& 0.8336& 0.8729& 0.6271\\ 
    \bottomrule
    \end{tabular}
    \caption{Ablation study of intent-aware mobility prediction model. `Full Model' refers to the complete structure, `Max-Prob' refers to uses the embedding of the most possible as the embedding, `Train-Real' refers to use the intent generated by LLM to train directly, `w/o intent' refers to structure without intent information}
    \vspace{-0.5cm}
    \label{tab:Ablation Study}
\end{table}

\section{Related Work} \label{sec:related}

\subsection{Mobility Prediction}
Mobility prediction involves anticipating future locations or visits of individuals based on historical mobility data. Before the population of deep learning based models, researchers employ Markov models~\cite{rendle2010factorizing, cheng2013you} to predict the next location by learning the transition relations between consecutive POIs. However, they fail to capture the high-order transitions among trajectories and also struggling in considering the effects of urban environment and user preference. Thus, research introduce the deep learning methods~\cite{liu2016predicting, feng2018deepmove, sun2020go, luo2021stan, yang2022getnext} to address these issues and achieve promising results. Liu et al.~\cite{liu2016predicting} first introduce RNN to predict the next location by proposing the spatial-temporal interpolation based ST-RNN. Feng et al.~\cite{feng2018deepmove} apply attention mechanism to capture the multi-scale temporal patterns of human mobility. Luo et al.~\cite{luo2021stan} further introduce the dual attention module to capture the spatial-temporal correlations among trajectories. While these methods succeed in modelling the sequential patterns in the trajectory, they cannot capture the shared mobility patterns between users effectively. To solve this problem, graph neural networks are introduced into the mobility prediction modelling. By representing trajectories as graph or hyper-graph~\cite{lim2020stp, yan2023spatio}, these approaches enable a comprehensive understanding of transition patterns at different locations~\cite{wang2023adaptive,yin2023next}, thus enriching the representations of users and POIs. However, due to the absence of large scale mobility intent dataset, these works ignore the intent modelling behind the mobility. In this work, we employ the agentic workflow to enable the large-scale automatic annotation of mobility intent and further propose an effective intent-aware mobility prediction model. 

\subsection{Large Language Models}
Large language models (LLMs)~\cite{achiam2023gpt, dubey2024llama} have achieved rapid development in the past few years, making significant progress in reasoning~\cite{wei2022cot}, planning, coding~\cite{roziere2023code}, mathematics~\cite{luo2023wizardmath}, and other fields. Some researchers also try to directly apply LLMs in the mobility modelling~\cite{wang2023would, beneduce2024large}. Different from these works, we use LLMs as the mobility data annotator to augment mobility data for training a stronger small domain-specific model. While LLMs are applied into many fields and show promising results to some extent, the lack of domain knowledge of LLMs on specific application limit their further improvement. Recently, LLM based agent framework~\cite{wang2024survey, xi2023rise} are proposed to complement the deficiencies of LLMs on specific domain knowledge and unleash the power of LLMs in real-world tasks, e.g.,  ChatDev~\cite{qian2023communicative} for project programming and WebAgent~\cite{gur2023real} for autonomous web tasks. Different from these works, we focus on the mobility prediction task and design the first agentic workflow for challenging intent annotation.

\section{Conclusion}
In this paper, we investigate the problem of intent enhanced mobility prediction problem. We propose LIMP, an agentic workflow based framework to harness the commonsense reasoning abilities of LLMs for intention-aware mobility prediction. We first design the "Analyze-Abstract-Infer" (A2I) framework to help the powerful LLM, GPT-4o in the paper, infer the mobility intent precisely. Then, we design agentic workflow fine-tuning mechanism to distilling common sense reasoning power of powerful LLM to smaller LLM, e.g., Llama3-8B. With the support of smaller LLM for large scale mobility data annotation, we finally design a transformer based model to complete the intent-aware mobility prediction task. Extensive experiments on two real-world datasets demonstrate the effectiveness of proposed framework. In the future, we plan to extend the framework to other spatial-temporal applications. 

\bibliography{custom}

\appendix
\section{Appendix} \label{sec:appendix}

Prompt example for generating high-level patterns of intent
\begin{lstlisting}
Your task is to extract the features of intent \'At Home\', \'Working\', \'Running errands\' from the statistical data. Please think step by step.
       
Here's the statistical data of the user's intent distribution:{{\'Intent 1\': {\'percentage_distribution\': 43.17, \'average_visit\': 2.5426470588235293}, \'Intent 2\': {\'percentage_distribution\': 15.23, \'average_visit\': 2.489795918367347}, \'Intent 3\': {\'percentage_distribution\': 25.89, \'average_visit\': 1.4730113636363635}, \'Time Distribution of Intents\': {\'Intent 1\': {0: 75.19, 1: 80.0, 2: 85.19, 3: 85.71, 4: 72.0, 5: 66.67, 6: 71.25, 7: 58.17, 8: 29.95, 9: 25.93, 10: 29.13, 11: 30.18, 12: 34.78, 13: 27.27, 14: 32.69, 15: 22.51, 16: 26.2, 17: 31.98, 18: 42.19, 19: 43.43, 20: 52.36, 21: 68.69, 22: 74.11, 23: 70.5}, \'Intent 2\': {0: 1.55, 1: 0.0, 2: 0.0, 3: 0.0, 4: 0.0, 5: 0.0, 6: 3.75, 7: 0.65, 8: 3.69, 9: 4.04, 10: 7.39, 11: 6.76, 12: 5.53, 13: 6.06, 14: 4.81, 15: 3.66, 16: 7.49, 17: 8.11, 18: 8.02, 19: 7.17, 20: 6.44, 21: 4.21, 22: 3.05, 23: 4.32}, \'Intent 3\': {0: 2.33, 1: 0.0, 2: 0.0, 3: 0.0, 4: 12.0, 5: 0.0, 6: 5.0, 7: 3.27, 8: 2.3, 9: 5.05, 10: 6.96, 11: 4.5, 12: 4.74, 13: 8.08, 14: 4.33, 15: 2.09, 16: 3.74, 17: 8.11, 18: 5.06, 19: 2.79, 20: 3.0, 21: 5.14, 22: 2.54, 23: 2.88}, \'Intent 4\': {0: 13.95, 1: 11.67, 2: 14.81, 3: 14.29, 4: 8.0, 5: 19.05, 6: 12.5, 7: 18.95, 8: 24.88, 9: 30.3, 10: 27.83, 11: 35.14, 12: 32.41, 13: 30.3, 14: 30.77, 15: 39.27, 16: 34.22, 17: 28.38, 18: 27.0, 19: 27.89, 20: 28.33, 21: 10.75, 22: 13.2, 23: 12.95}, \'Intent 5\': {0: 3.1, 1: 5.0, 2: 0.0, 3: 0.0, 4: 0.0, 5: 0.0, 6: 0.0, 7: 5.88, 8: 11.06, 9: 5.72, 10: 6.96, 11: 7.21, 12: 9.49, 13: 7.58, 14: 6.25, 15: 8.9, 16: 10.7, 17: 7.66, 18: 6.75, 19: 5.58, 20: 3.43, 21: 4.21, 22: 1.02, 23: 4.32}, \'Intent 6\': {0: 3.88, 1: 3.33, 2: 0.0, 3: 0.0, 4: 8.0, 5: 14.29, 6: 7.5, 7: 13.07, 8: 28.11, 9: 28.96, 10: 21.74, 11: 16.22, 12: 13.04, 13: 20.71, 14: 21.15, 15: 23.56, 16: 17.65, 17: 15.77, 18: 10.97, 19: 13.15, 20: 6.44, 21: 7.01, 22: 6.09, 23: 5.04}}}}

The meanings of statistical data are as follows:
- Percentage Distribution: The percent of the intent in the whole dataset.
- Time Distribution: The start_time distribution of visits to the POI with the intent, in the format of (start hour: percentage).

There are 6 intents in total: ['At Home', 'Working', 'Running errands', 'Eating Out', 'Leisure and entertainment', 'Shopping'], each intent has a percentage distribution and a time distribution.

Instruction:
- You need to extract the unique and prominent features of intent 'At Home', 'Working', 'Running errands' which can distinguish them from other intents.
- Each intent should have about 6-8 features.
- Should be based on the percentage distribution, and time distribution of the intent.
- Should be able to help identify the user's home, work place, and running errands place.
- Some features need to be specificity to the intent, such as the time distribution of the intent.

Answer using the following JSON format:
{{
"features": ["features of 'intents'"],
}}
\end{lstlisting} 

Prompt example for home and work identification.
\begin{lstlisting}
Your objective is to identify the potential 'home,' and 'work' places of a user's intent based on their trajectory data and the features associated with the intents 'At Home' and 'Working'. Please think step by step.
                           
The trajectory data under analysis is as follows: {[{\'Name\': \'poi1\', \'Percent\': \'8.5%
        
The meaning of each element in the trajectory data is as follows:
    - Name: the POI the user visited.
    - Percent: The percentage of times the behavior pattern occurred
    - Time Distribution: the start time distribution of the number of visits to the POI, in the format of (start hour: percentage).

Here are the general and unique features of intent 'At Home' , 'Working' , 'Running errands':{            "intent": "At Home",
            "features": [
                "High percentage distribution: 43.17%
                "Average visit: 2.54 times",
                "Peak time distribution: Early morning (0-3 AM) and late evening (8-11 PM)",
                "Consistent presence throughout the day with notable dips during typical working hours (8 AM - 5 PM)",
                "Significant presence during late night hours (12 AM - 3 AM)",
                "Gradual increase in presence from 4 PM onwards, peaking at 9 PM",
                "High presence during night hours (10 PM - 12 AM)"
            ]
        },
        {
            "intent": "Working",
            "features": [
                "Moderate percentage distribution: 15.23%
                "Average visit: 2.49 times",
                "Peak time distribution: Morning (8-11 AM) and early afternoon (12-3 PM)",
                "Significant presence during typical working hours (8 AM - 5 PM)",
                "Notable drop in presence during early morning (0-6 AM) and late evening (6 PM onwards)",
                "Presence peaks at 9 AM and 10 AM, indicating start of workday",
                "Gradual decrease in presence after 3 PM"
            ]
        },
        {
            "intent": "Running errands",
            "features": [
                "Moderate percentage distribution: 25.89%
                "Average visit: 1.47 times",
                "Peak time distribution: Late morning (9-11 AM) and early afternoon (12-3 PM)",
                "Significant presence during mid-morning to early afternoon (9 AM - 3 PM)",
                "Notable drop in presence during early morning (0-6 AM) and late evening (6 PM onwards)",
                "Presence peaks at 11 AM, indicating common time for running errands",
                "High presence during mid-day hours (10 AM - 2 PM)"
            ]
        }
    ]}

Respond using the following JSON format:
{{
    "home": "home place",
    "work": "work place"
    "reason": "reason for prediction"
}}
\end{lstlisting} 

Prompt example for intent prediction.
\begin{lstlisting}
Your task is to give intent prediction using trajectory data. Let's think step by step.

1. Analyze the user's behavior pattern based on the trajectory data.
2. Consider and think about the name of the POI and the time distribution of visits to the POI with the intent. (This is the trajectory of one person, so thinking about the user's daily routine is important.)
3. Based on the user's behavior pattern and please consider the features of intent 'At Home', 'Working', 'Running errands', predict the intent of each stay in the trajectory data.

The trajectory data under analysis is as follows: {(poi name1, 2019-10-11 00:30:00, 2019-10-11 07:30:00)(poi name2, 2019-10-11 08:15:00, 2019-10-11 15:30:00)(poi name3, 2019-10-11 15:45:00, 2019-10-11 17:00:00)}.

Each stay in trajectory data is represented as (poi, start time).

Here's what each element means:
- poi: the POI the user visited.
- start time: the time the user arrived at the POI.

Please judge the function of POI based on its name, time distribution, and features provided. You should take the meaning of each intent as reference, but the final judgment shouldn't be fully rely on that.

Intent you can choose:['At Home', 'Working', 'Running errands', 'Eating Out', 'Leisure and entertainment', 'Shopping']

Here's what each intent means:
- At Home: When the user is at {poi name1}, it is mostly considered as being at home. And Other places are NOT considered as home! 
- Working: When the user is at {poi name2}, it is mostly considered as working. And Other places are NOT considered as working!
But, you should still consider the user\'s behavior pattern, POI_name, and the time the user visited the POI.

Note: If multiple conditions are met, priority should be given to 'At Home' and 'Running Errands'.

There are {3} stays in the trajectory data. So, the output should have {3} predicted intents.

Consider step by step, finally respond using the following JSON format (Make sure to have one predicted intent for each stay in the trajectory data, And you have to assign one of the intents to each stay in the trajectory data):
{{
"predicted_intent": ["adjusted predicted intents"],
}}
\end{lstlisting}

Prompt example for zero-shot intent prediction.
\begin{lstlisting}
Your task is to give intent prediction using trajectory data. 
        
The trajectory data under analysis is as follows: {(poi name1, 2019-10-11 00:30:00, 2019-10-11 07:30:00)(poi name2, 2019-10-11 08:15:00, 2019-10-11 15:30:00)(poi name3, 2019-10-11 15:45:00, 2019-10-11 17:00:00)}.

Each stay in trajectory data is represented as (poi, start time).

Here's what each element means:
    - POI: the POI the user visited.
    - Start Time: the time the user arrived at the POI.

Intent you can choose:['At Home', 'Working', 'Running errands', 'Eating Out', 'Leisure and entertainment', 'Shopping']

There are {3} stays in the trajectory data. So make sure the output should only have {3} predicted intents.

Respond using the following JSON format to provide the predicted intents:
{{
"predicted_intent": ["adjusted predicted intents"],
}}
\end{lstlisting} 

Prompt example of home and work identification with feature ablation
\begin{lstlisting}
Your objective is to identify the potential 'home,' and 'work' places of a user's intent based on their trajectory data and the features associated with the intents 'At Home' and 'Working'. Please think step by step.
               
The trajectory data under analysis is as follows: {[{\'Name\': \'poi1\', \'Percent\': \'8.5%

The meaning of each element in the trajectory data is as follows:
- Name: the POI the user visited.
- Percent: The percentage of times the behavior pattern occurred
- Time Distribution: the start time distribution of number of visits to the POI, in the format of (start hour: percentage).

Respond using the following JSON format:
{{
"home": "home place",
"work": "work place",
"reason": "reason for prediction"
}}
\end{lstlisting}

Prompt example for intent prediction with features ablation 
\begin{lstlisting}
Your task is to give intent prediction using trajectory data. Let's think step by step.

1. Analyze the user's behavior pattern based on the trajectory data.
2. Consider and think about the name of the POI and the time distribution of visits to the POI with the intent. (This is the trajectory of one person, so thinking about the user's daily routine is important.)
3. Based on the user's behavior pattern and please consider the features of intent 'At Home', 'Working', 'Running errands', predict the intent of each stay in the trajectory data.

The trajectory data under analysis is as follows: {(poi name1, 2019-10-11 00:30:00, 2019-10-11 07:30:00)(poi name2, 2019-10-11 08:15:00, 2019-10-11 15:30:00)(poi name3, 2019-10-11 15:45:00, 2019-10-11 17:00:00)}.

Each stay in trajectory data is represented as (poi, start time).

Here's what each element means:
- poi: the POI the user visited.
- start time: the time the user arrived at the POI.

Please judge the function of POI based on its name, time distribution, and features provided. You should take the meaning of each intent as reference, but the final judgment shouldn't be fully rely on that.

Intent you can choose:['At Home', 'Working', 'Running errands', 'Eating Out', 'Leisure and entertainment', 'Shopping']

Here's what each intent means:
- At Home: When the user is at {poi name1}, it is mostly considered as being at home. And Other places are NOT considered as home! 
- Working: When the user is at {poi name2}, it is mostly considered as working. And Other places are NOT considered as working!
But, you should still consider the user\'s behavior pattern, POI_name, and the time the user visited the POI.

Note: If multiple conditions are met, priority should be given to 'At Home' and 'Running Errands'.

There are {} stays in the trajectory data. So, the output should have {} predicted intents.

Consider step by step, finally respond using the following JSON format (Make sure to have one predicted intent for each stay in the trajectory data, And you have to assign one of the intents to each stay in the trajectory data):
{{
"predicted_intent": ["adjusted predicted intents"],
}}
\end{lstlisting}

Prompt and answer example for fine-tune task 1.
\begin{lstlisting}
Your task is to identify the user's home and work place based on the trajectory data and the features of intent 'At Home' and 'Working'.
The trajectory data under analysis is as follows: [{'Name': 'poi name1', 'Percent': '79.2%
Each entry represents a POI-intent pair that the user has visited.\n    The meanings of each feature are as follows:
- Name: POI name
- Percent: The percentage of times the behavior pattern occurred
- Time Distribution: The time distribution of visits to the POI with the intent, in the format of (hour, percentage).
Here are the features of intent 'At Home' and 'Working':{'features': [{'intent': 'At Home', 'features': ['Highest overall percentage distribution at 43.2%
{\"home\": \"home place\",\"work\": \"work place\"}

answer:
{"home": "poi name1", "work": "poi name2"}

\end{lstlisting}

Prompt and answer example for fine-tune task 2.
\begin{lstlisting}
"Your task is to give intent prediction using trajectory data. Stay in trajectory data corresponds one by one to intent.
The trajectory data under analysis is as follows: (poi name1, High School, 2019-11-18 01:00:00)(poi name1, High School, 2019-11-18 13:15:00)(poi name2, Educational Facilities, 2019-11-19 00:00:00).
Each stay in trajectory data is represented as (poi, category of poi, start time).
Here's what each element means:\n    - poi: the POI the user visited.
- category of poi: category the POI belongs to.
- start time: the time the user arrived at the POI.
Please mainly judge the function of POI based on its name. The POI category can be used to assist in judgment.
Intent you can choose:['At Home', 'Working', 'Running errands', 'Eating Out', 'Leisure and entertainment', 'Shopping']
Here's what each intent means:
- At Home: When the user is at poi name2, it is always considered as being at home, regardless of the time and POI category. When the user is at other places, it is not considered as being at home.
- Working: When the user is at poi name1, it is always considered as working, regardless of the time and POI category. When the user is at other places, it is not considered as working.
- Running errands: When the user is not at poi name2 or poi name1, and the POI is unlikely to be a place for shopping, entertainment or eating, it is considered as running errands.
Note: If multiple conditions are met, priority should be given to 'At Home' and 'Working'.\n    There are 3 stays in the trajectory data. So, the output should have 3 predicted intents.
Respond using a list: ["intent1", "intent2" ...]"

answer:
['Working','Working','At Home']

\end{lstlisting}

\end{document}